\pdfoutput=1

\documentclass[11pt]{article}

\usepackage[final]{acl}

\usepackage{times}
\usepackage{latexsym}

\usepackage[T1]{fontenc}

\usepackage[utf8]{inputenc}

\usepackage{microtype}

\usepackage{inconsolata}
\usepackage{array}
\usepackage{ragged2e}
\usepackage{graphicx}
\usepackage{makecell}
\usepackage{hhline}
\usepackage{hyperref}
\usepackage{amsmath}
\usepackage{enumitem}
\usepackage{bm}
\newcommand{\changeA}[1]{\textcolor{black}{#1}}

\title{Scalable and Domain-General Abstractive Proposition Segmentation}

\author{Mohammad Javad Hosseini$^{1}$ \quad Yang Gao$^{1}$ \quad Tim Baumgärtner$^{2}$\thanks{Work done as an intern at Google.} \\ \quad {\bf Alex Fabrikant}$^{1}$ \quad {\bf Reinald Kim Amplayo}$^{1}$ \\
        $^{1}$Google DeepMind \\ $^{2}$Ubiquitous Knowledge Processing Lab, Technical University of Darmstadt \\
        \texttt{\{javadh,gaostayyang,fabrikant,reinald\}@google.com} \\
        \texttt{tim.baumgaertner@tu-darmstadt.de}
}

\begin{document}
\maketitle
\begin{abstract}

Segmenting text into fine-grained units of meaning is important to a wide range of NLP applications. The default approach of segmenting text into sentences is often insufficient, especially since sentences are usually complex enough to include multiple units of meaning that merit separate treatment in the downstream task. We focus on the task of \textit{abstractive proposition segmentation} (APS): transforming text into simple, self-contained, well-formed sentences. Several recent works have demonstrated the utility of proposition segmentation with few-shot prompted LLMs for downstream tasks such as retrieval-augmented grounding and fact verification. However, this approach does not scale to large amounts of text and may not always extract all the facts from the input text.
In this paper, we first introduce evaluation metrics for the task to measure several dimensions of quality.
We then propose a scalable, yet accurate, proposition segmentation model. We model proposition segmentation as a supervised task by training LLMs on existing annotated datasets and show that training yields significantly improved results. We further show that by using the fine-tuned LLMs (Gemini Pro and Gemini Ultra) as teachers for annotating large amounts of multi-domain synthetic distillation data, we can train smaller student models (Gemma 1 2B and 7B) with results similar to the teacher LLMs. We then demonstrate that our technique leads to effective domain generalization, by annotating data in two domains outside the original training data and evaluating on them. Finally, as a key contribution of the paper, we share an easy-to-use API\footnote{\changeA{Our  \href{https://huggingface.co/collections/google/gemma-aps-release-66e1a42c7b9c3bd67a0ade88}{Gemma-APS} API (Gemma 1 2B and 7B) can be found on Hugging Face.}} for NLP practitioners to use.

\end{abstract}

\section{Introduction}

From retrieval systems that build indices over passages rather than documents \cite{tiedemann2008simple}, to automatic evaluation metrics for
generative tasks that evaluate sentence-level similarity to references
(e.g. \citet{amplayo2022smart}), to structured event representations used for
cross-document summarization \cite{zhang-etal-2023-enhancing}, segmenting a document into significantly finer units that retain relevant meaning is a major component of many NLP systems.

In ``well-formed'' prose, an easy and frequently used choice for
segmenting documents is sentence segmentation. But for most
applications, sentences are an imperfect fit: they are often still too
complex, containing multiple units of underlying
information \cite{chen2023propsegment, min2023factscore}; they typically require context from
elsewhere in the document to understand the meaning \cite{choi2021decontextualization}.
Furthermore, well-formed sentences are not always available in situations ranging from casual speech \cite{stainton2005defense} to non-prose formats \cite{fang2024large, maheshwari2024presentations}, where ``sentences'' are not even a natural unit of discourse.

To provide useful fine-grained segmentation, several recent works have taken the approach of \textit{proposition segmentation}\footnote{Others in the literature have also used terms such as ``claim decomposition'', ``claim extraction'', and ``atomic fact extraction'' for the same concept. We follow the naming in \cite{chen2023propsegment}.} \cite{chen2023propsegment, min2023factscore, wanner2024closer}, seeking to break text into fine-grained, minimal units of meaning that together convey all the information in the source text. Similarly to the extractive-vs-abstractive contrast in the summarization literature, the two strands of proposition segmentation work so far have considered either (a) an \textit{extractive} approach, representing propositions as one or more spans in the source text (\citet{chen2023propsegment}; \citet{gunel2023strum}; etc.); or (b) few-shot LLM prompts for \textit{abstractive} proposition segmentation, generatively writing out each unit as a well-formed sentence \cite{kamoi-etal-2023-wice, wanner2024closer,scire-etal-2024-fenice}.

More formally, abstractive proposition segmentation (APS), the focus of this paper, is to transform a given document into a collection of \textit{propositions} represented as natural-language sentences which: 1. are atomic and minimal semantic unit that cannot be further decomposed into meaningful units \cite{liu-etal-2023-revisiting}; 2. are fully decontextualized \cite{choi2021decontextualization} --- i.e. they can be understood just as well with no access to the rest of the document; 3. present information explicitly given in the document; and 4. when taken together, cover all of the information in the document.

APS has already found applications in grounding \cite{gao-etal-2023-rarr},  summarization evaluation \cite{liu-etal-2023-revisiting,scire-etal-2024-fenice}, and fact checking \cite{min2023factscore}. In this paper, we focus on \textit{making abstractive proposition segmentation practical}. The few-shot prompting approaches are typically too costly to run at large scales, and, furthermore, we show that they tend to under-extract compared to our proposed solutions.
Our core contributions are:

\begin{enumerate}[itemsep=-1ex]
\item \textbf{A suite of automatic evaluation metrics} to measure the quality of APS methods along several relevant dimensions, allowing informed comparisons between methods

\item \textbf{Supervision by existing datasets \cite{liu-etal-2023-revisiting}}, which empirically shows improvement on APS over few-shot prompting baselines.

\item \textbf{Scalable, domain-general student models (Gemma 1 2B and 7B, \citet{team2024gemma})} for APS distilled from the supervised models over synthetic multi-domain data \cite{hosseini2024synthetic}, yielding performance comparable to the teacher models even on domains not seen in the human-annotated training data.

\item \textbf{An APS API} for NLP practitioners to use.
\end{enumerate}

\section{Related Work}

Linguistic compositionality, the idea that sentences are comprised by smaller units of meaning, has been debated since the early 1800s \cite{janssen-compositionality}, and understood surely long beforehand. In the context of modern NLP, the value of proposition segmentation for standard tasks can be seen from the empirical measurements in, e.g., \cite{chen2023propsegment}, which shows for document-level NLI that 72\% of sentences partially aligned between two highly related documents don't fully entail each other, and in \cite{min2023factscore}, which shows that 40\% of ChatGPT sentences at that time contained a mix of supported and unsupported propositions.

Indeed, several previous results have shown APS by few-shot prompted LLMs benefits retrieval-augmented fact verification and grounding \cite{kamoi-etal-2023-wice, min2023factscore}. A concurrent result in \cite{wanner2024closer} looks more specifically at APS itself with few-shot prompting. \changeA{\citet{scire-etal-2024-fenice} also perform few-shot prompting followed by distillation.}

Other formats of proposition segmentation have also been explored. Extractive proposition segmentation is shown in \cite{chen2023propsegment, chen2023dense} to benefit document-level NLI and retrieval. Several open-book QA and grounding works have generated \textit{fine-grained questions} corresponding implicitly to the fine-grained claims in the text \cite{gao-etal-2023-rarr, chen-etal-2022-generating, chen2023complex}.
In the summarization evaluation literature, ``Summary Content Units'', initially human-annotated \cite{nenkova-passonneau-2004-evaluating-pyramid}, later generated from syntactic signals \cite{gao-etal-2019-automated} have long been used for summary evaluation. Before the LLM era, decomposing text into semantic triples, known as Open Information Extraction \cite{etzioni2008open}, drove a variety of downstream applications.

Our desiderata for proposition segmentation include context-independence, earlier studied at the sentence level by  \citet{choi2021decontextualization}. \changeA{\citet{deng-etal-2024-document} perform document-level claim extraction for fact checking. They specifically extract claims that are check-worthy, where these claims are decontextualized, but not necessarily atomic.}

\changeA{In our work, we propose a suite of automatic evaluation metrics. Previous efforts have not focused much on metric definition. The exceptions are two concurrent works: A) \citet{wanner2024closer} propose a specific single metric for APS, DecompScore, that combines our reference-free precision metric (\S\ref{sec:metrics}) with the count of claims generated. B) \citet{scire-etal-2024-fenice} propose metrics based on ROUGE \cite{lin2004rouge} and similar to our reference-based precision and recall. Our metrics are based on NLI that is more suitable for checking semantic equivalence of predicted and gold propositions.}

\section{Abstractive Proposition Segmentation}
\label{sec:problem_def}

In this section, we formally define the task (\S\ref{sec:task}) and propose metrics (\S\ref{sec:metrics}).

\subsection{Task Definition}
\label{sec:task}

We are given an input text $t$, which comprises a naturally-occurring sequence of English words,
possibly split into multiple sentences, i.e., $t=\{s_1,...,s_n\}$.
In text $t$, there are $k$ latent \changeA{gold propositions (claims)} $\{p_1,...,p_k\}$.
The task is then to segment $t$ into a list of propositions $\{q_1,...,q_k\}$ with the following conditions:

\begin{enumerate}[itemsep=-1ex]
    \item \textbf{Well-formed}: Proposition $q_i$ should be grammatically correct and conform to the rules of the English language.
    \item \textbf{Atomic}: Proposition $q_i$ should contain a single atomic fact.
    \item \textbf{Self-contained}: Proposition $q_i$ should not need additional context to be understood.
    \item \textbf{Supported}: Proposition $q_i$ should be found in the given text $t$.
    \item \textbf{Comprehensive}:
    The list of propositions $\{q_1,...,q_k\}$ should cover all the  \changeA{latent gold propositions (claims)} in text $t$.\looseness=-1
\end{enumerate}

\subsection{Evaluation Metrics}
\label{sec:metrics}

To evaluate systems that produce propositions following the conditions above, we propose two sets of metrics that make use of an entailment model.
We employ Natural Language Inference (NLI) as 
backbone to our metrics because by definition \cite{dagan2022recognizing}, it can be used to check factual support (i.e., one entails another) and semantic equivalence (i.e., both entail each other). \changeA{In addition, NLI has been successfully used for evaluating factual consistency \cite{team2023gemini,gao-etal-2023-enabling,fierro-etal-2024-learning,honovich-etal-2022-true}.}

We use a T5-11B model \cite{raffel2020exploring} fine-tuned on the ANLI dataset \cite{nie2020adversarial} as our entailment model $\texttt{NLI}(\texttt{premise}, \texttt{claim})$ that returns an entailment score between 0 and 1. \changeA{ This model is shown to obtain the highest factual consistency results on the TRUE benchmark in \citet{honovich-etal-2022-true}'s experiments.}

\paragraph{Reference-free (RF)} The first set of metrics compare the system-generated propositions $Q=\{q_1,...,q_{k'}\}$ with input text $t=\{s_1,...,s_n\}$, which helps us evaluate whether the propositions are supported and comprehensive. Specifically, we calculate precision $RF_p$ and recall $RF_r$ as follows:

{\small
\setlength{\abovedisplayskip}{6pt}
\setlength{\belowdisplayskip}{\abovedisplayskip}
\setlength{\abovedisplayshortskip}{0pt}
\setlength{\belowdisplayshortskip}{3pt}
\begin{align}
    RF_p &= \frac{\sum_{q_i \in Q} \texttt{NLI}(\texttt{premise}=t, \texttt{claim}=q_i)}{k} \\
    RF_r &= \frac{\sum_{s_j \in t} \texttt{NLI}(\texttt{premise}=\bar{Q}, \texttt{claim}=s_j)}{n} 
\end{align}
}

where $\bar{Q}$ is the space-concatenated version of ${Q}$ to create a single text. Here, precision essentially
evaluates whether each proposition $q_i$ in $Q$ is supported in text $t$, while recall evaluates whether each latent gold proposition mentioned
in each sentence $s_j$ is covered in $Q$. We can then combine both precision and recall by calculating the f1-score $RF_{f1}$.

\paragraph{Reference-based (RB)} The second set of metrics rely on a gold-standard set of propositions $P=\{p_1,...,p_k\}$ and check whether each proposition in $P$ is semantically equivalent to a predicted proposition (and vice versa). To this end, we use a bidirectional version of NLI where $\texttt{premise}$ and $\texttt{claim}$ need to entail each other, i.e.:

{\small
\setlength{\abovedisplayskip}{6pt}
\setlength{\belowdisplayskip}{\abovedisplayskip}
\setlength{\abovedisplayshortskip}{0pt}
\setlength{\belowdisplayshortskip}{3pt}
\begin{equation}
\texttt{BiNLI}(p_i, q_j) = \texttt{min} \big(\texttt{NLI}(p_i, q_j), \texttt{NLI}(q_j, p_i)\big)
\end{equation}
}

\changeA{The first NLI call (i.e., does gold entail predicted?) ensures atomicity: If the predicted proposition $q_j$ is not as atomic as a gold proposition $p_i$, then $p_i$ will not entail $q_j$ (since $q_j$ has more information that $p_i$).}
\changeA{The second NLI call (i.e., does predicted entail gold?) ensures self-containedness:  If the predicted proposition $q_j$ is not as self-contained as a gold proposition $p_i$, then $q_j$ will not entail $p_i$ (since $p_i$ has more information than $q_j$).}

$q_j$ should not need further context (otherwise, the entailment does not hold).
We calculate precision $RB_p$ and recall $RB_r$ as follows:

{\small
\setlength{\abovedisplayskip}{6pt}
\setlength{\belowdisplayskip}{\abovedisplayskip}
\setlength{\abovedisplayshortskip}{0pt}
\setlength{\belowdisplayshortskip}{3pt}
\begin{align}
    RB_p &= \frac{\sum_{q_j \in Q} \texttt{argmax}_{p_i \in P} \texttt{BiNLI}(p_i, q_j)}{k'} \\
    RB_r &= \frac{\sum_{p_i \in P} \texttt{argmax}_{q_j \in Q} \texttt{BiNLI}(p_i, q_j)}{k}
\end{align}
}

In $RB_p$ metric, for each predicted $q_j$, we find the most equivalent $p_i$ based on $\texttt{BiNLI}(p_i, q_j)$, and then average over all predicted propositions. $RB_r$ is calculated similarly in the other direction.
Finally, we can combine both precision and recall by calculating the f1-score $RB_{f1}$. We note that our reference-based scores are equivalent to SMART metrics proposed by \newcite{amplayo2022smart} as an evaluation metric for text generation. They treat sentences as basic units of information, and compare the set of gold and predicted sentences.
We compare propositions as basic units of information rather than sentences.

Note that we do not measure
well-formedness since we assume such property for system predictions, given the advancements of pretrained LMs.

\changeA{In order to validate the effectiveness of our metrics, we perform human correlation studies and show positive results (\S\ref{sec:human-correlation}). We show examples of how metrics are calculated in Appendix \ref{sec:metric_calc_examples}.}

\section{Domain-General APS}

Given an input text (passage) $t$,  our goal is to generate a list of propositions $\{p_1,\ldots,p_k\}$, where propositions should be well-formed, atomic, self-contained, supported, and comprehensive.

In this section, we discuss our proposed method to distill a relatively small, yet domain general proposition segmentation model: A) We train a teacher LLM on an existing proposition segmentation dataset (\S\ref{sec:finetune}). B) We generate a large set of multi-domain synthetic data with different lengths (\S\ref{sec:syn-data}). C) We generate a large synthetic dataset with pairs of (text, propositions list) and train a student model on it (\S\ref{sec:distillation}).

\subsection{Training an APS Model}
\label{sec:finetune}

\begin{figure*}[ht!]
    \centering
    \includegraphics[width=15cm]{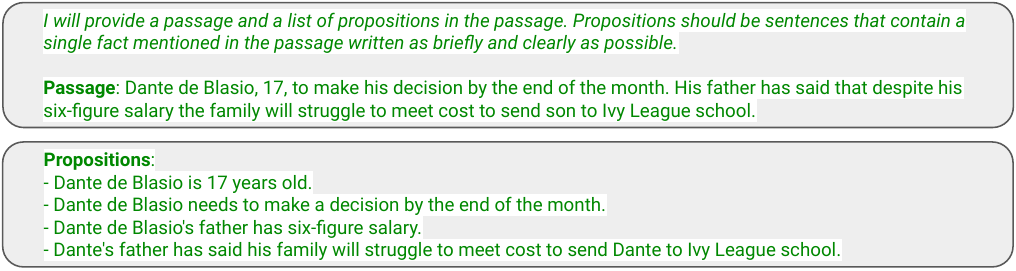}
    \caption{The input (top) and output (bottom) for training an APS model with ungrouped propositions. The input contains an instruction and a passage. The output contains the list of propositions.}
    \label{fig:inp_out1}
    \vspace{-4mm}
\end{figure*}

\begin{figure*}[ht!]
    \centering
    \includegraphics[width=15cm]{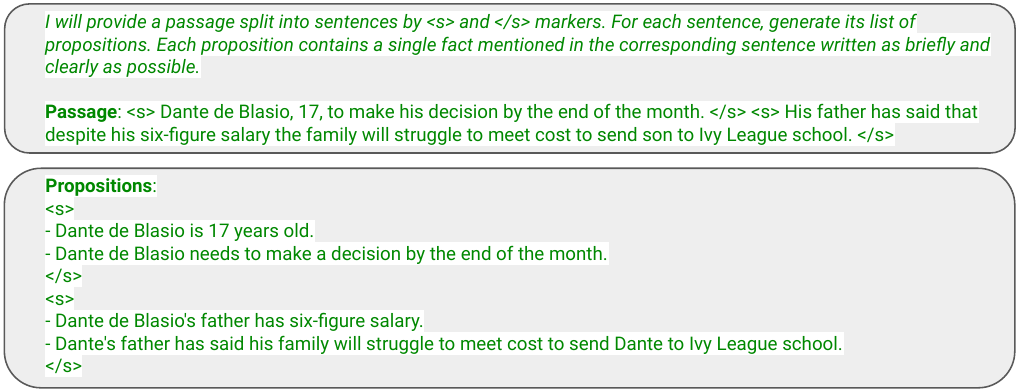}
    \caption{The input (top) and output (bottom) for training an APS model with grouped propositions. The input contains an instruction and a passage. The output contains the list of propositions. The input passage is separated by special start and end of sentence tokens. Similarly, the output propositions of each sentence are grouped together using special tokens.}
    \label{fig:inp_out2}
    \vspace{-4mm}
\end{figure*}

We train a teacher APS model based on an LLM.
In particular, we train a model by using examples in the ROSE dataset \cite{liu-etal-2023-revisiting}. Each example contains an input text $t$, and a list of propositions $\{p_1,\ldots,p_k\}$.
We trained using two approaches: ungrouped propositions and grouped propositions.

In the ungrouped propositions version, the input contains an instruction and a passage (Figure \ref{fig:inp_out1} top). We add an instruction since we use instruction-tuned LLMs for training.
The output contains the list of propositions each prepended by ``-'' and separated by a newline character (Figure \ref{fig:inp_out1} bottom).

In the grouped propositions version, we leverage the existing sentence structure from the passage. We split the passage into sentences before feeding it into the proposition segmentation model. We specify the sentence boundaries with special start of sentence (<s>) and end of sentence (</s>) tokens. In addition, we group the propositions of each sentence together and place them inside start and end of sentence tokens. Figure \ref{fig:inp_out2} shows an example.

The grouped propositions approach has two benefits:
A) The trained model could use the sentence boundaries to obtain improved performance, since it can learn how to generate propositions per sentence rather than generating a longer list of propositions for the full passage. B) During inference, we can automatically attribute each proposition to its corresponding sentence. This is useful for downstream applications. For example, in grounding applications, we can spot which sentences have propositions that are supported or contradicted by an arbitrary source.

We fine-tuned two different LLMs as our teachers: Gemini Pro
and Gemini Ultra \cite{team2023gemini}.\footnote{Available from \href{https://cloud.google.com/apis}{https://cloud.google.com/apis}}

\subsection{Generating Multi-Domain Synthetic Data}
\label{sec:syn-data}

In order to generate a synthetic dataset for distillation, we require a large set of passages so that we can apply the teacher LLM to them and produce (text, propositions) pairs. The ROSE dataset contains examples only in the news domain. To have maximum generalization to new domains, the passages should cover as many domains as possible. In addition, the passage should have different lengths so that the model works well with new texts of different lengths.

We follow \newcite{hosseini2024synthetic} that take a practical approach and consider various text properties as contributing factors to domains: text genre, topic, and even the platform or venue that the text comes from. They design a prompt with $18$ few-shot examples, where each example is a triple of (\textit{length}, \textit{domain}, \textit{text}). The length can take either the value \textit{short} (just one or a few sentences) or \textit{paragraph}. Appendix \ref{sec:few-shot-example} shows an example.\footnote{The full list can be found in \newcite{hosseini2024synthetic}.}
The set of $18$ few-shot examples cover $8$ seed domains such as \textit{shopping reviews}, \textit{twitter} and \textit{reddit post}.
However, to have a wide range of domains, they first prompt FLAN-PaLM2 L (Unicorn) model \cite{palm2} to generate new domains. Then, they manually select a number of non-repetitive domains. Finally, they prompt the LLM to generate text in those domains with the two lengths.

We replicated their approach using Gemini Ultra \cite{team2023gemini}. We first prompted Gemini Ultra $4,000$ times to generate new domains.\footnote{\changeA{Many of the calls generate one of the existing domains from the few-shot examples. Therefore, in order to obtain many unseen domains, we prompted the LLM $4,000$ times.}} We obtained a set of $105$ domains, from which we manually selected $75$. We then prompted the LLM and generated $226$K examples with the selected domains and the two lengths.\footnote{We first generated $228$K examples, but filtered examples with $n\geq4$-gram overlap with any of the seed examples ($\approx2$K examples).}

\subsection{Distillation}
\label{sec:distillation}
The teacher proposition segmentation LLMs learn the task well since they have a large number of parameters and are supervised trained on the ROSE dataset (\S\ref{sec:finetune}). However, they are too costly for direct use in practical applications. Therefore, we distill them into student models.

In our preliminary experiments, we observed better results from the grouped propositions version (\S\ref{sec:exp-in-domain}), so we trained the student model based on this type of teacher. We apply the teacher LLMs to the synthetic multi-domain set of texts (\S\ref{sec:syn-data}) and produce $226$K (text, propositions) pairs.  We then train a model with the same input and output format as the teacher models with grouped proposition.
We used Gemma 1\footnote{\changeA{We used Gemma 1 in all our experiments, but we refer to the language model as Gemma for simplicity throughout most of the following text.}} (2B and 7B) \cite{team2024gemma}, a lightweight state-of-the-art LM\footnote{\href{https://ai.google.dev/gemma}{https://ai.google.dev/gemma}}, as our student model.

\section{Experiments}
\label{sec:exp}

We explore the effectiveness of our distillation approach for training a \textit{scalable} and \textit{domain-general} proposition segmentation approach. We describe the datasets we have used for training and evaluation (\S\ref{sec:datasets}). We then introduce our baselines (\S\ref{sec:baselines}).  We first compare our proposed method with multiple baselines on the ROSE dataset (\S\ref{sec:exp-in-domain}). We then show that our method is effective on two datasets from new and unseen domains (\S\ref{sec:exp-out-of-domain}).

\subsection{Datasets}
\label{sec:datasets}

We use the annotated ROSE dataset for supervised training. The ROSE dataset has examples from the \textit{news} domain. We manually annotate two out-of-domain datasets, ensuring that the propositions have the desired properties (\S\ref{sec:problem_def}). We use these datasets for assessing the domain generalization capabilities of our models.

\textbf{ROSE dataset}. This dataset is built by manually splitting news summaries into Atomic Content Units (ACUs) for the purpose of evaluating such summaries \cite{liu-etal-2023-revisiting}. \changeA{The ACUs are annotated based on a set of well-defined rules to extract atomic facts, i.e., elementary information units in the input text which no longer need to be further split for the purpose of ambiguity reduction for human evaluation \cite{liu-etal-2023-revisiting}.}

The ACU definition in the ROSE dataset are very close to our propositions definition, therefore we used them for training. We observed some cases in the dataset where the propositions are either not supported or not comprehensive, but we filtered those examples automatically.
The dataset contains $2,500$ passages ($21,797$ propositions). We randomly split the dataset into a training and development set (for hyper-parameter tuning\footnote{See details of hyper-parameters in Appendix \ref{sec:hparam}.}). The training set contains $2,089$ passages ($18,994$ propositions), and the development set contains $411$ passages ($2,803$ propositions).

The original dataset contains the full set of propositions for each passage. However, for training the grouped propositions version (\S\ref{sec:finetune}), we need to align each proposition to a sentence. We pre-process the dataset to obtain such alignment. We use the NLI score (from T5 11B trained on ANLI) between sentences (premise) and propositions (hypothesis) to obtain the alignment. In particular, for each proposition, we find the sentence with the maximum NLI score. If the NLI score from that sentence $\geq \tau=0.9$, we align the proposition to the sentence. Otherwise, we discard the example (unsupported proposition). After aligning all the propositions, if a sentence is not aligned with any proposition, we again discard the example (a special case of non-comprehensive propositions). We provide more details about the alignment, filtering, and pre-processing in Appendix \ref{sec:alignment_and_filter}.

The final dataset has high $RL_p$ (supported) and $RL_r$ (comprehensive) scores (\S\ref{sec:exp-in-domain}). We manually evaluated the alignment on $\approx 200$ propositions from the ROSE development set, and the error rate of this approach was $\approx 2\%$. The final training and development sets contain $1,923$ examples ($15,092$ propositions) and $383$ examples ($2,237$ propositions), respectively.

\changeA{Since the examples in the ROSE dataset are based on news summaries, they are quite general and cover many linguistic forms such as presuppositions, attribution to the speaker, modals, and sentence connectors (e.g., \textit{because} and \textit{however}).}

\textbf{Reddit}. The Reddit dataset contains $20$ randomly sampled human-written answer passages from WebGPT \cite{nakano2021webgpt}, which is a subset of ELI5 dataset, originally used for long-form question answering \cite{fan-etal-2019-eli5}. We sampled from one paragraph long answers. We manually annotated the passages with propositions.

\textbf{Amazon Review}. The Amazon Review dataset contains $20$ randomly sampled reviews with $3$ to $7$ sentences from the 2018 version\footnote{\url{https://cseweb.ucsd.edu/~jmcauley/datasets/amazon_v2/}} of the Amazon Review Data \cite{ni-etal-2019-justifying}. We specifically sampled from the 5-core subset. Finally, we manually annotated each review with propositions.

\changeA{The manual annotations for the Reddit and Amazon Review datasets were done by two of the authors (each annotated one dataset). The instructions were based on the task definition (\S\ref{sec:task}). The authors looked at examples from the ROSE dataset to be mostly aligned with those examples as well.}

\subsection{Baselines}
\label{sec:baselines}

\begin{table*}[ht!] 
\centering 
\small 
\begin{tabular}{|m{0.15\linewidth}|*{3}{m{0.06\linewidth}}|*{3}{m{0.06\linewidth}}|m{0.06\linewidth}|} 
& \multicolumn{3}{c}{\bf \textsc{Reference-Less Metrics}} & \multicolumn{3}{c}{\bf \textsc{Reference-Based Metrics}} & \\& Precision & Recall & F1 & Precision & Recall & F1 & \# Props \\
\hline
Gold & 99.71 & 96.54 & 97.53 & 100.00 & 100.00 & 100.00 & 5.84 \\ \hline
Sentence & 100.00 & 100.00 & 100.00 & 24.71 & 18.24 & 20.42 & 2.52 \\ \hline
\multicolumn{8}{|c|}{ {\bf \textsc{ Few Shot } } } \\ \hline
Gemini Pro Dyn & 99.21 & 93.26 & 94.31 & 47.10 & 41.41 & 43.20 & 4.22 \\ \hline
Gemini Ultra Dyn & 99.37 & 89.75 & 91.88 & 49.49 & 47.49 & 47.74 & 5.29 \\ \hline
\multicolumn{8}{|c|}{ {\bf \textsc{ Trained on ROSE } } } \\ \hline
Gemma 7B UG & 98.09 & 96.57 & 96.54 & 52.16 & 50.93 & 51.02 & 5.57 \\ \hline
Gemma 7B G & 98.57 & 97.48 & 97.48 & 53.70 & 51.43 & 51.93 & 5.61 \\ \hline
Gemini Pro UG & 99.51 & 97.84 & 98.20 & 54.76 & 52.48 & 53.02 & 5.54 \\ \hline
Gemini Pro G & 99.31 & 96.66 & 97.23 & 55.96 & 54.87 & 54.83 & 5.66 \\ \hline
Gemini Ultra UG & 99.46 & 98.05 & 98.33 & \textbf{57.69} & 56.32 & 56.45 & 5.72 \\ \hline
Gemini Ultra G & \textbf{99.53} & \textbf{98.16} & \textbf{98.50} & 57.62 & \textbf{56.50} & \textbf{56.49} & \textbf{5.77} \\ \hline
\multicolumn{8}{|c|}{ {\bf \textsc{ Gemma 7B Distilled Models } } } \\ \hline
Gemini Pro Data & 98.98 & 97.91 & 98.02 & 55.14 & 53.02 & 53.50 & 5.53 \\ \hline
Gemini Ultra Data & 98.93 & 98.08 & 98.23 & 56.82 & 55.18 & 55.41 & 5.65 \\ \hline
\end{tabular}
\caption{Results on the \textsc{ROSE} dataset. The methods are split into $4$ blocks. The first block has gold and sentence baselines. The second one has few-shot baselines with dynamically (Dyn) selected examples. The third block has baselines directly trained on ROSE with ungrouped (UG) and grouped (G) propositions. The fourth block contains the distilled models results. The best result for each metric (excluding gold and sentence baselines) is boldfaced.}
\label{tab:res_rose}
\end{table*}

We compare the following set of models:

{\bf Gold} has the human annotated propositions.

{\bf Sentence} is a trivial baseline where we consider each sentence as a proposition.

{\bf Few Shot} extracts propositions by few-shot prompting an LLM. For each test example, we selected the most similar $K=10$ examples from the training set based on ROUGE-1 score \cite{lin2004rouge}. We report the results for two LLMs, Gemini Pro and Gemini Ultra. We also tried two additional few-shot prompting approaches: A) Randomly sampling few-shot examples (average of $5$ runs), B) The few-shot examples from \cite{wanner2024closer}. In both cases, the results were overall worse than the dynamic approach and had a similar pattern compared to our models (Appendeix \ref{sec:full-results}).

{\bf Trained on ROSE} are cases where we supervised trained a LM. We trained two versions for each language model (\S\ref{sec:finetune}): ungrouped propositions (UG) and grouped propositions (G). We trained Gemma 7B, Gemini Pro and Gemini Ultra. We also tried Gemma 2B and obtained consistent results with 7B (Appendix \ref{sec:full-results}). Moreover, we did preliminary experiments with T5 and obtained consistent results, although lower than Gemma.

{\bf Gemma 7B Distilled Models} are our final models (\S\ref{sec:distillation}). We fine-tuned Gemma 7B as the student model on distillation data from Gemini Pro and Ultra teacher models (grouped propositions).

\begin{table*}[ht] 
\centering 
\small 
\begin{tabular}{|m{0.15\linewidth}|*{3}{m{0.06\linewidth}}|*{3}{m{0.06\linewidth}}|m{0.06\linewidth}|} 
& \multicolumn{3}{c}{\bf \textsc{Reference-Less Metrics}} & \multicolumn{3}{c}{\bf \textsc{Reference-Based Metrics}} & \\& Precision & Recall & F1 & Precision & Recall & F1 & \# Props \\
\hline
Gold & 98.72 & 99.22 & 98.86 & 100.00 & 100.00 & 100.00 & 10.70 \\ \hline
Sentence & 100.00 & 100.00 & 100.00 & 32.99 & 17.80 & 22.43 & 4.40 \\ \hline
\multicolumn{8}{|c|}{ {\bf \textsc{ Few Shot } } } \\ \hline
Gemini Pro Dyn & \textbf{100.00} & 83.40 & 89.31 & \textbf{56.98} & 42.03 & 47.74 & 7.30 \\ \hline
Gemini Ultra Dyn & 98.99 & 72.61 & 80.44 & 54.59 & 44.73 & \textbf{48.53} & 8.15 \\ \hline
\multicolumn{8}{|c|}{ {\bf \textsc{ Trained on ROSE } } } \\ \hline
Gemma 7B G & 98.25 & 98.73 & 98.35 & 35.49 & 38.30 & 36.57 & 10.25 \\ \hline
Gemini Pro G & 98.80 & 94.41 & 96.08 & 41.80 & 43.44 & 42.06 & \textbf{10.65} \\ \hline
Gemini Ultra G & 99.68 & 96.66 & 97.83 & 40.82 & 44.69 & 42.39 & 11.20 \\ \hline
\multicolumn{8}{|c|}{ {\bf \textsc{ Gemma 7B Distilled Models } } } \\ \hline
Gemini Pro Data & 99.47 & 97.08 & 98.00 & 45.22 & \textbf{48.08} & 46.20 & 10.90 \\ \hline
Gemini Ultra Data & 98.88 & \textbf{99.96} & \textbf{99.40} & 40.43 & 43.21 & 41.46 & 11.00 \\ \hline
\end{tabular}
\caption{Results on the \textsc{REDDIT} dataset. See Table \ref{tab:res_rose}'s caption for details.}
\label{tab:res_reddit}
\end{table*}

\begin{table*}[ht] 
\centering 
\small 
\begin{tabular}{|m{0.15\linewidth}|*{3}{m{0.06\linewidth}}|*{3}{m{0.06\linewidth}}|m{0.06\linewidth}|} 
& \multicolumn{3}{c}{\bf \textsc{Reference-Less Metrics}} & \multicolumn{3}{c}{\bf \textsc{Reference-Based Metrics}} & \\& Precision & Recall & F1 & Precision & Recall & F1 & \# Props \\
\hline
Gold & 100.00 & 99.98 & 99.99 & 100.00 & 100.00 & 100.00 & 6.55 \\ \hline
Sentence & 100.00 & 100.00 & 100.00 & 37.94 & 24.50 & 29.21 & 3.55 \\ \hline
\multicolumn{8}{|c|}{ {\bf \textsc{ Few Shot } } } \\ \hline
Gemini Pro Dyn & 99.54 & 62.97 & 71.82 & 50.02 & 46.74 & 47.89 & 6.10 \\ \hline
Gemini Ultra Dyn & 85.09 & 53.10 & 57.17 & 44.41 & 44.38 & 42.06 & 11.60 \\ \hline
\multicolumn{8}{|c|}{ {\bf \textsc{ Trained on ROSE } } } \\ \hline
Gemma 7B G & \textbf{99.98} & \textbf{99.99} & \textbf{99.99} & 51.90 & 48.88 & 49.83 & 6.25 \\ \hline
Gemini Pro G & 99.53 & 97.98 & 98.50 & 55.62 & 54.09 & 54.52 & 6.60 \\ \hline
Gemini Ultra G & 99.83 & 96.98 & 98.09 & \textbf{56.75} & 57.08 & \textbf{56.65} & 6.80 \\ \hline
\multicolumn{8}{|c|}{ {\bf \textsc{ Gemma 7B Distilled Models } } } \\ \hline
Gemini Pro Data & 98.30 & 96.72 & 97.30 & 56.00 & \textbf{57.09} & 56.25 & 7.05 \\ \hline
Gemini Ultra Data & 99.27 & 99.16 & 99.21 & 53.43 & 53.03 & 53.09 & \textbf{6.55} \\ \hline
\end{tabular}
\caption{Results on the \textsc{AMAZON REVIEW} dataset. See Table \ref{tab:res_rose}'s caption for details.}
\label{tab:res_amazon}
\end{table*}

\subsection{In-Domain Results}
\label{sec:exp-in-domain}

We first compare our method with all the baselines on ROSE development set. Table \ref{tab:res_rose} shows the results.
We split the metrics (columns) into two main blocks: reference-less and reference-based (\S\ref{sec:metrics}). In addition, we report the average number of propositions per baseline. In the ideal scenario, the average number of predicted propositions should be as close as possible to gold propositions.

\textit{\textbf{Gold and sentence baselines}}. The gold propositions have very high $RL_p$ ($99.71$\%) and $RL_r$ ($96.54$\%), which shows that the pre-processed dataset (\S\ref{sec:datasets}) has high quality and satisfy the supported and comprehensive conditions.
The $RB$ metrics, on the other hand, are $100\%$ by definition. The average number of propositions is $5.84\%$. The sentence baseline has perfect $RL$ metrics by definition. However, the $RB$ metrics are very low.

\textit{\textbf{Few-shot models}}. These baselines (Gemini Pro Dyn and Gemini Ultra Dyn) have very high $RL_{p}$ ($99.21$\% and $99.37\%$), but their $RL_r$ ($93.26$\% and $89.75$\%) is relatively low compared to supervised baselines. The $RB$ metrics are considerably lower than trained models.

\textit{\textbf{Grouped vs Ungrouped versions}}. Among the trained models, we observe that the grouped ones outperform the ungrouped ones (with only a few exceptions). For examples, Gemma 7B G has $97.48\%$ $RL_{f1}$ ($51.93\%$ $RB_{f1}$), while the UG version has $96.54\%$ $RL_{f1}$ ($51.02\%$ $RB_{f1}$). Therefore, we performed distillation data generation (\S\ref{sec:distillation}) using the grouped propositions version. We also note that the grouped propositions trained models always output the correct format in our experiments, i.e., they output an equal number of start and end tokens, and the same number of groups as the sentences.

\textit{\textbf{Size of trained LMs}}. Larger LMs get better results than smaller ones when trained on ROSE (with only a few exceptions): Gemini Ultra gets better results compared to Gemini Pro, which itself gets better results than Gemma 7B.

\textit{\textbf{Student models}}.
We trained two different Gemma 7B student models, one trained on distillation data from Gemini Pro teacher model, and one from Gemini Ultra teacher model. The Gemma 7B student models outperform Gemma 7B trained directly on ROSE (\changeA{i.e., with no distillation}) on all metrics. In addition, Gemma 7B student models (the last two rows) perform close to their corresponding teacher models (the last two rows of the trained on ROSE block).

\textit{\textbf{Number of predicted propositions}}. The number of predicted propositions correlate well with $RB$ metrics and follow similar patterns.

\subsection{Out-of-Domain Results}
\label{sec:exp-out-of-domain}

Table \ref{tab:res_reddit} and Table \ref{tab:res_amazon} show the results of different models on the Reddit and Amazon Review datasets.

\textit{\textbf{Gold and sentence baselines}}. The gold data has high $RL$ metrics confirming that our annotations satisfy supported and comprehensive conditions. The sentence baselines have perfect $RL$ metrics by definition, but very low $RB$ metrics.

\textit{\textbf{Student models vs teacher models and training directly on ROSE}}. In both datasets, all the trained and distilled models have very high $RL$ metrics ($RL_{f1} \geq 96\%$). However, the student models perform significantly better than Gemma 7B trained directly on ROSE (\changeA{i.e., with no distillation}) in $RB$ metrics. This confirms that our distillation approach using synthetic multi-domain data leads to successful domain adaptation.
In addition, the student models get results on par with teacher models (and sometimes even better) on out-of-domain datasets.

\begin{table*}[ht!]
\small
\centering
\begin{tabular}{| m{3.5cm} | m{4.7cm} | m{4.7cm} | m{1.2cm} |}
\hline
\textbf{Input text} & \textbf{Gold Propositions} & \textbf{Predicted Propositions} & \textbf{Category} \\
\hline
There are lots of things that feel good that carry some kind of risk. & \textbf{There are lots of things that feel good that carry some kind of risk.} & - \textbf{There are lots of things that feel good.} - \textbf{There are lots of things that carry some kind of risk.} & Atomicity \\ \hline
Fits well and is stylish! Light weight and great options such as the stand. Cant beat this one for the money. & - It fits well. - It is stylish! - It is light weight. - \textbf{It has great options.} - \textbf{One great option of it is the stand.} - Cant beat it for the money. & - Fits well. - Is stylish. - Light weight. \textbf{Great options such as the stand.} - Cant beat this one for the money. & Atomicity \\ \hline
I've always used this type of pick for playing bass. I like the material and the thickness is just right. & - I've always used this type of pick for playing bass. - \textbf{I like the material of this type of pick.} - \textbf{The thickness of this type of pick is just right.} & - I've always used this type of pick for playing bass. - \textbf{I like the material.} - \textbf{the thickness is just right.} & Decont \\ \hline
But fish near reefs (or in small streams) have other options. They have shelter to hide behind if they spot a predator (meaning camouflage isn't as important). & - Fish near reefs have other options. - Fish in small streams have other options. - \textbf{Fish near reefs have shelter to hide behind if they spot a predator.} - \textbf{Fish in small streams have shelter to hide behind if they spot a predator.} - \textbf{Camouflage isn't as important as hiding behind shelters.} & Fish near reefs have other options. - Fish in small streams have other options. - \textbf{Fish have shelter to hide behind.} - \textbf{Fish can hide behind if they spot a predator.} - \textbf{Camouflage isn't as important.} & Decont \\ \hline
\end{tabular}
\caption{Examples where gold propositions and predicted propositions are not paraphrase because they do not have the exact atomicity or decontextualization level (boldfaced propositions). However, the predicted propositions are not necessarily wrong especially when it comes to the atomicity level. In these cases, the $RB$ scores will be $0$.}
\label{tab:gold_vs_predicted}
\end{table*}

\textit{\textbf{Few-shot models compared to student and teacher}}.
The few-shot models have very low $RL_r$ ($53\%$ to $83\%$) compared to the student models ($\geq 97$\%). This makes the few-shot models unreliable for downstream applications such as fact verification that require to have access to all the claims in the input passage. Table \ref{tab:fewshot_vs_student} in Appendix \ref{sec:fewshot_vs_student} shows examples. The $RB$ metrics for few-shot models is slightly better than the student models on Reddit, but much worse on Amazon Review.

\textit{\textbf{Note on $\bm{RB}$ metrics}}. The $RB$ metrics are very strict when comparing gold and predicted propositions, and some minor changes from the gold propositions could lead to low $RB$ metrics. In particular, when computing $RB_p$, if a predicted proposition is not a paraphrase of any gold proposition, then it will have a score $=0$ (\S\ref{sec:metrics}).

In many cases, it is challenging and subjective to decide on the right level of atomicity and decontextualization, which directly affects $RB$ metrics (\S\ref{sec:metrics}). Table \ref{tab:gold_vs_predicted} shows a number of examples where our annotated and predicted propositions (Gemma 7B distilled from Gemini Pro data) are different, although the predicted ones are not necessarily wrong especially when it comes to the atomicity level. For example, the sentence ``There  are  lots  of  things  that feel good that carry some kind of risk'' has the right level of atomicity to be considered as a proposition if we want to emphasize on the two points jointly (``feeling good'' and ``carrying some kind of risk''). Otherwise, the sentence could be split into two propositions. In our work, since we trained the teacher LLMs on the ROSE dataset, we expect the trained models to mirror the atomicity levels in the dataset.
\section{\changeA{Human Correlation Studies for APS Metrics}}
\label{sec:human-correlation}
We measure how well our defined metrics (\S\ref{sec:metrics}) align with human judgements in order to validate the metrics' effectiveness. We performed a study on $40$ passages ($142$ sentences, $263$ predicted propositions, and $262$ gold propositions) from the Amazon Review dataset. We used the predictions from two models: Gemini Pro few-shot and Gemma 7B distilled from fine-tuned Gemini Pro. The annotations were done by two of the authors (each example were annotated by one author). For each input sentence, we annotated whether the predicted propositions cover all the claims in the sentence (used for measuring reference-less recall). For each predicted proposition, we annotated whether it is supported by the input passage (reference-less precision) and whether it is equivalent to any of the gold propositions (reference-based precision). For each gold proposition, we annotated whether it is equivalent to any of the predicted propositions (reference-based recall).

The Pearson correlation coefficients of example-level metrics and human annotations were generally high (Table \ref{tab:pearson_correlation}), confirming that our proposed metrics do correlate well with human judgements (p-value < $0.01$).\footnote{The reference-less precision metric is almost always equal to 1 except for a few examples. The NLI accuracy compared to human annotation is $0.985$. In addition, both the automatic metric and the human evaluated metric are $>=0.98$.}

\begin{table}[ht!]
\centering
\small 
\begin{tabular}{|m{0.41\linewidth}|m{0.45\linewidth}|}
\hline
\textbf{Metric}                 & \textbf{Pearson Correlation Coefficient} \\ \hline
Reference-based Pr        & 0.718 \\ \hline
Reference-based Rec           & 0.731 \\ \hline
Reference-less Pr         & 0.476 \\ \hline
Reference-less Rec            & 0.647 \\ \hline
\end{tabular}
\caption{Pearson correlation coefficients of metrics and human judgements (p-value < $0.01$).}
\label{tab:pearson_correlation}
\end{table}

\section{The \texttt{propositions} API}

We showed that our student models resolve two issues with the commonly used few-shot prompting approach: under-extraction (low $RL_r$) and cost.
As part of this paper, we release the \href{https://huggingface.co/collections/google/gemma-aps-release-66e1a42c7b9c3bd67a0ade88}{Gemma-APS} API on Hugging Face based on Gemma 1 2B G and Gemma 1 7B G student models trained from Gemini Pro data (grouped propositions version). We invite researchers that require proposition segmentation on input text to try out our models instead of few-shot prompting LLMs.

\section{Conclusion}

We define the abstractive proposition segmentation task more formally by specifying the desired properties of propositions and present a suite of automatic evaluation metrics that allow us to measure different dimensions of quality. While previous work often uses few-shot prompting, we show that supervision from existing datasets yields significant quality improvement. We then propose a distillation approach for training scalable and domain-general models that get on-par results with the teachers (and sometimes even better). We release an API based on Gemma 7B student models and invite researchers to use that instead of few-shot prompting LLMs.

\section{Limitations}

In our analysis we showed that reference-based metrics depend on the atomicity and decontextualization level of propositions. On the other hand, the right level of atomicity and decontextualization depends on the downstream applications and how propositions will be used. In addition, our models outputs mirror the atomicity and decontextualization levels of the ROSE dataset examples. Future models and metrics could be flexible in these two levels and let the user decide on the actual style needed for their downstream application.

\changeA{We used NLI as the backbone to our metrics. We note that NLI as a task is not fully solved, and there are some levels of disagreement in human annotation \cite{pavlick-kwiatkowski-2019-inherent,weber-genzel-etal-2024-varierr}. However, we showed strong correlations between human judgements and the defined metrics. In addition, NLI computation is done using a fine-tuned language model, so it is not very lightweight. However, the metric computation usually needs to be done on a small dataset.}

We performed our experiments only on English; however, our abstractive proposition segmentation definition and proposed metrics are language independent. In addition, we observed multilingual capabilities with the teacher models when tried on examples from multiple languages. This capability could be used for training multilingual student models in the future.

We note that although our proposition segmentation model is quite accurate and outperforms existing approaches, it is still possible for it to generate wrong and hallucinated outputs, as with all other baselines. Downstream applications should be attuned to the possibility of APS outputs that are occasionally not supported by the original documents.

\bibliography{custom}

\begin{thebibliography}{42}
\providecommand{\natexlab}[1]{#1}

\bibitem[{Amplayo et~al.(2023)Amplayo, Liu, Zhao, and
  Narayan}]{amplayo2022smart}
Reinald~Kim Amplayo, Peter~J. Liu, Yao Zhao, and Shashi Narayan. 2023.
\newblock \href {https://openreview.net/pdf?id=OIe3kpwl40D} {{SMART:} sentences
  as basic units for text evaluation}.
\newblock In \emph{The Eleventh International Conference on Learning
  Representations, {ICLR} 2023, Kigali, Rwanda, May 1-5, 2023}. OpenReview.net.

\bibitem[{Anil et~al.(2023{\natexlab{a}})Anil, Borgeaud, Wu, Alayrac, Yu,
  Soricut, Schalkwyk, Dai, Hauth, Millican, Silver, Petrov, Johnson,
  Antonoglou, Schrittwieser, Glaese, Chen, Pitler, Lillicrap, Lazaridou, Firat,
  Molloy, Isard, Barham, Hennigan, Lee, Viola, Reynolds, Xu, Doherty, Collins,
  Meyer, Rutherford, Moreira, Ayoub, Goel, Tucker, Piqueras, Krikun, Barr,
  Savinov, Danihelka, Roelofs, White, Andreassen, von Glehn, Yagati, Kazemi,
  Gonzalez, Khalman, Sygnowski, and et~al.}]{team2023gemini}
Rohan Anil, Sebastian Borgeaud, Yonghui Wu, Jean{-}Baptiste Alayrac, Jiahui Yu,
  Radu Soricut, Johan Schalkwyk, Andrew~M. Dai, Anja Hauth, Katie Millican,
  David Silver, Slav Petrov, Melvin Johnson, Ioannis Antonoglou, Julian
  Schrittwieser, Amelia Glaese, Jilin Chen, Emily Pitler, Timothy~P. Lillicrap,
  Angeliki Lazaridou, Orhan Firat, James Molloy, Michael Isard, Paul~Ronald
  Barham, Tom Hennigan, Benjamin Lee, Fabio Viola, Malcolm Reynolds, Yuanzhong
  Xu, Ryan Doherty, Eli Collins, Clemens Meyer, Eliza Rutherford, Erica
  Moreira, Kareem Ayoub, Megha Goel, George Tucker, Enrique Piqueras, Maxim
  Krikun, Iain Barr, Nikolay Savinov, Ivo Danihelka, Becca Roelofs,
  Ana{\"{\i}}s White, Anders Andreassen, Tamara von Glehn, Lakshman Yagati,
  Mehran Kazemi, Lucas Gonzalez, Misha Khalman, Jakub Sygnowski, and et~al.
  2023{\natexlab{a}}.
\newblock \href {https://doi.org/10.48550/ARXIV.2312.11805} {Gemini: {A} family
  of highly capable multimodal models}.
\newblock \emph{CoRR}, abs/2312.11805.

\bibitem[{Anil et~al.(2023{\natexlab{b}})Anil, Dai, Firat, Johnson, Lepikhin,
  Passos, Shakeri, Taropa, Bailey, Chen, Chu, Clark, Shafey, Huang,
  Meier{-}Hellstern, Mishra, Moreira, Omernick, Robinson, Ruder, Tay, Xiao, Xu,
  Zhang, {\'{A}}brego, Ahn, Austin, Barham, Botha, Bradbury, Brahma, Brooks,
  Catasta, Cheng, Cherry, Choquette{-}Choo, Chowdhery, Crepy, Dave, Dehghani,
  Dev, Devlin, D{\'{\i}}az, Du, Dyer, Feinberg, Feng, Fienber, Freitag, Garcia,
  Gehrmann, Gonzalez, and et~al.}]{palm2}
Rohan Anil, Andrew~M. Dai, Orhan Firat, Melvin Johnson, Dmitry Lepikhin,
  Alexandre Passos, Siamak Shakeri, Emanuel Taropa, Paige Bailey, Zhifeng Chen,
  Eric Chu, Jonathan~H. Clark, Laurent~El Shafey, Yanping Huang, Kathy
  Meier{-}Hellstern, Gaurav Mishra, Erica Moreira, Mark Omernick, Kevin
  Robinson, Sebastian Ruder, Yi~Tay, Kefan Xiao, Yuanzhong Xu, Yujing Zhang,
  Gustavo~Hern{\'{a}}ndez {\'{A}}brego, Junwhan Ahn, Jacob Austin, Paul Barham,
  Jan~A. Botha, James Bradbury, Siddhartha Brahma, Kevin Brooks, Michele
  Catasta, Yong Cheng, Colin Cherry, Christopher~A. Choquette{-}Choo, Aakanksha
  Chowdhery, Cl{\'{e}}ment Crepy, Shachi Dave, Mostafa Dehghani, Sunipa Dev,
  Jacob Devlin, Mark D{\'{\i}}az, Nan Du, Ethan Dyer, Vladimir Feinberg,
  Fangxiaoyu Feng, Vlad Fienber, Markus Freitag, Xavier Garcia, Sebastian
  Gehrmann, Lucas Gonzalez, and et~al. 2023{\natexlab{b}}.
\newblock \href {https://doi.org/10.48550/ARXIV.2305.10403} {Palm 2 technical
  report}.
\newblock \emph{CoRR}, arXiv:2305.10403.

\bibitem[{Chen et~al.(2023{\natexlab{a}})Chen, Kim, Sriram, Durrett, and
  Choi}]{chen2023complex}
Jifan Chen, Grace Kim, Aniruddh Sriram, Greg Durrett, and Eunsol Choi.
  2023{\natexlab{a}}.
\newblock \href {https://doi.org/10.48550/ARXIV.2305.11859} {Complex claim
  verification with evidence retrieved in the wild}.
\newblock \emph{CoRR}, abs/2305.11859.

\bibitem[{Chen et~al.(2022)Chen, Sriram, Choi, and
  Durrett}]{chen-etal-2022-generating}
Jifan Chen, Aniruddh Sriram, Eunsol Choi, and Greg Durrett. 2022.
\newblock \href {https://doi.org/10.18653/v1/2022.emnlp-main.229} {Generating
  literal and implied subquestions to fact-check complex claims}.
\newblock In \emph{Proceedings of the 2022 Conference on Empirical Methods in
  Natural Language Processing}, pages 3495--3516, Abu Dhabi, United Arab
  Emirates. Association for Computational Linguistics.

\bibitem[{Chen et~al.(2023{\natexlab{b}})Chen, Buthpitiya, Fabrikant, Roth, and
  Schuster}]{chen2023propsegment}
Sihao Chen, Senaka Buthpitiya, Alex Fabrikant, Dan Roth, and Tal Schuster.
  2023{\natexlab{b}}.
\newblock \href {https://doi.org/10.18653/v1/2023.findings-acl.565}
  {{P}rop{S}egm{E}nt: A large-scale corpus for proposition-level segmentation
  and entailment recognition}.
\newblock In \emph{Findings of the Association for Computational Linguistics:
  ACL 2023}, pages 8874--8893, Toronto, Canada. Association for Computational
  Linguistics.

\bibitem[{Chen et~al.(2023{\natexlab{c}})Chen, Wang, Chen, Yu, Ma, Zhao, Zhang,
  and Yu}]{chen2023dense}
Tong Chen, Hongwei Wang, Sihao Chen, Wenhao Yu, Kaixin Ma, Xinran Zhao,
  Hongming Zhang, and Dong Yu. 2023{\natexlab{c}}.
\newblock \href {https://doi.org/10.48550/ARXIV.2312.06648} {Dense {X}
  retrieval: What retrieval granularity should we use?}
\newblock \emph{CoRR}, abs/2312.06648.

\bibitem[{Choi et~al.(2021)Choi, Palomaki, Lamm, Kwiatkowski, Das, and
  Collins}]{choi2021decontextualization}
Eunsol Choi, Jennimaria Palomaki, Matthew Lamm, Tom Kwiatkowski, Dipanjan Das,
  and Michael Collins. 2021.
\newblock \href {https://doi.org/10.1162/tacl_a_00377} {Decontextualization:
  Making sentences stand-alone}.
\newblock \emph{Transactions of the Association for Computational Linguistics},
  9:447--461.

\bibitem[{Dagan et~al.(2013)Dagan, Roth, Sammons, and
  Zanzotto}]{dagan2022recognizing}
Ido Dagan, Dan Roth, Mark Sammons, and Fabio~Massimo Zanzotto. 2013.
\newblock \href {https://doi.org/10.2200/S00509ED1V01Y201305HLT023}
  {\emph{Recognizing Textual Entailment: Models and Applications}}.
\newblock Synthesis Lectures on Human Language Technologies. Morgan {\&}
  Claypool Publishers.

\bibitem[{Davidson(1967)}]{davidson1967logical}
Donald Davidson. 1967.
\newblock The logical form of action sentences.
\newblock In Nicholas Rescher, editor, \emph{The Logic of Decision and Action},
  pages 81--95. University of Pittsburgh Press.

\bibitem[{Deng et~al.(2024)Deng, Schlichtkrull, and
  Vlachos}]{deng-etal-2024-document}
Zhenyun Deng, Michael Schlichtkrull, and Andreas Vlachos. 2024.
\newblock \href {https://doi.org/10.18653/v1/2024.acl-long.645} {Document-level
  claim extraction and decontextualisation for fact-checking}.
\newblock In \emph{Proceedings of the 62nd Annual Meeting of the Association
  for Computational Linguistics (Volume 1: Long Papers)}, pages 11943--11954,
  Bangkok, Thailand. Association for Computational Linguistics.

\bibitem[{Etzioni et~al.(2008)Etzioni, Banko, Soderland, and
  Weld}]{etzioni2008open}
Oren Etzioni, Michele Banko, Stephen Soderland, and Daniel~S. Weld. 2008.
\newblock \href {https://doi.org/10.1145/1409360.1409378} {Open information
  extraction from the web}.
\newblock \emph{Commun. ACM}, 51(12):68–74.

\bibitem[{Fan et~al.(2019)Fan, Jernite, Perez, Grangier, Weston, and
  Auli}]{fan-etal-2019-eli5}
Angela Fan, Yacine Jernite, Ethan Perez, David Grangier, Jason Weston, and
  Michael Auli. 2019.
\newblock \href {https://doi.org/10.18653/v1/P19-1346} {{ELI}5: Long form
  question answering}.
\newblock In \emph{Proceedings of the 57th Annual Meeting of the Association
  for Computational Linguistics}, pages 3558--3567, Florence, Italy.
  Association for Computational Linguistics.

\bibitem[{Fang et~al.(2024)Fang, Xu, Tan, Zhang, Hu, Qi, Nickleach, Socolinsky,
  Sengamedu, and Faloutsos}]{fang2024large}
Xi~Fang, Weijie Xu, Fiona~Anting Tan, Jiani Zhang, Ziqing Hu, Yanjun Qi, Scott
  Nickleach, Diego Socolinsky, Srinivasan~H. Sengamedu, and Christos Faloutsos.
  2024.
\newblock \href {https://doi.org/10.48550/ARXIV.2402.17944} {Large language
  models(llms) on tabular data: Prediction, generation, and understanding - {A}
  survey}.
\newblock \emph{CoRR}, arXiv:2402.17944.

\bibitem[{Fierro et~al.(2024)Fierro, Amplayo, Huot, De~Cao, Maynez, Narayan,
  and Lapata}]{fierro-etal-2024-learning}
Constanza Fierro, Reinald~Kim Amplayo, Fantine Huot, Nicola De~Cao, Joshua
  Maynez, Shashi Narayan, and Mirella Lapata. 2024.
\newblock \href {https://doi.org/10.18653/v1/2024.acl-long.615} {Learning to
  plan and generate text with citations}.
\newblock In \emph{Proceedings of the 62nd Annual Meeting of the Association
  for Computational Linguistics (Volume 1: Long Papers)}, pages 11397--11417,
  Bangkok, Thailand. Association for Computational Linguistics.

\bibitem[{Gao et~al.(2023{\natexlab{a}})Gao, Dai, Pasupat, Chen, Chaganty, Fan,
  Zhao, Lao, Lee, Juan, and Guu}]{gao-etal-2023-rarr}
Luyu Gao, Zhuyun Dai, Panupong Pasupat, Anthony Chen, Arun~Tejasvi Chaganty,
  Yicheng Fan, Vincent Zhao, Ni~Lao, Hongrae Lee, Da-Cheng Juan, and Kelvin
  Guu. 2023{\natexlab{a}}.
\newblock \href {https://doi.org/10.18653/v1/2023.acl-long.910} {{RARR}:
  Researching and revising what language models say, using language models}.
\newblock In \emph{Proceedings of the 61st Annual Meeting of the Association
  for Computational Linguistics (Volume 1: Long Papers)}, pages 16477--16508,
  Toronto, Canada. Association for Computational Linguistics.

\bibitem[{Gao et~al.(2023{\natexlab{b}})Gao, Yen, Yu, and
  Chen}]{gao-etal-2023-enabling}
Tianyu Gao, Howard Yen, Jiatong Yu, and Danqi Chen. 2023{\natexlab{b}}.
\newblock \href {https://doi.org/10.18653/v1/2023.emnlp-main.398} {Enabling
  large language models to generate text with citations}.
\newblock In \emph{Proceedings of the 2023 Conference on Empirical Methods in
  Natural Language Processing}, pages 6465--6488, Singapore. Association for
  Computational Linguistics.

\bibitem[{Gao et~al.(2019)Gao, Sun, and Passonneau}]{gao-etal-2019-automated}
Yanjun Gao, Chen Sun, and Rebecca~J. Passonneau. 2019.
\newblock \href {https://doi.org/10.18653/v1/K19-1038} {Automated pyramid
  summarization evaluation}.
\newblock In \emph{Proceedings of the 23rd Conference on Computational Natural
  Language Learning (CoNLL)}, pages 404--418, Hong Kong, China. Association for
  Computational Linguistics.

\bibitem[{Gunel et~al.(2023)Gunel, Tata, and Najork}]{gunel2023strum}
Beliz Gunel, Sandeep Tata, and Marc Najork. 2023.
\newblock \href {https://doi.org/10.1145/3543873.3587304} {{STRUM:} extractive
  aspect-based contrastive summarization}.
\newblock In \emph{Companion Proceedings of the {ACM} Web Conference 2023,
  {WWW} 2023, Austin, TX, USA, 30 April 2023 - 4 May 2023}, pages 28--31.
  {ACM}.

\bibitem[{Honovich et~al.(2022)Honovich, Aharoni, Herzig, Taitelbaum,
  Kukliansy, Cohen, Scialom, Szpektor, Hassidim, and
  Matias}]{honovich-etal-2022-true}
Or~Honovich, Roee Aharoni, Jonathan Herzig, Hagai Taitelbaum, Doron Kukliansy,
  Vered Cohen, Thomas Scialom, Idan Szpektor, Avinatan Hassidim, and Yossi
  Matias. 2022.
\newblock \href {https://doi.org/10.18653/v1/2022.dialdoc-1.19} {{TRUE}:
  Re-evaluating factual consistency evaluation}.
\newblock In \emph{Proceedings of the Second DialDoc Workshop on
  Document-grounded Dialogue and Conversational Question Answering}, pages
  161--175, Dublin, Ireland. Association for Computational Linguistics.

\bibitem[{Hosseini et~al.(2024)Hosseini, Petrov, Fabrikant, and
  Louis}]{hosseini2024synthetic}
Mohammad~Javad Hosseini, Andrey Petrov, Alex Fabrikant, and Annie Louis. 2024.
\newblock \href {https://doi.org/10.48550/ARXIV.2402.12368} {A synthetic data
  approach for domain generalization of {NLI} models}.
\newblock \emph{CoRR}, abs/2402.12368.

\bibitem[{Janssen(2012)}]{janssen-compositionality}
Theo Janssen. 2012.
\newblock \href {https://doi.org/10.1093/oxfordhb/9780199541072.013.0001} {{19
  Compositionality: Its Historic Context}}.
\newblock In \emph{{The Oxford Handbook of Compositionality}}. Oxford
  University Press.

\bibitem[{Kamoi et~al.(2023)Kamoi, Goyal, Diego~Rodriguez, and
  Durrett}]{kamoi-etal-2023-wice}
Ryo Kamoi, Tanya Goyal, Juan Diego~Rodriguez, and Greg Durrett. 2023.
\newblock \href {https://doi.org/10.18653/v1/2023.emnlp-main.470} {{W}i{CE}:
  Real-world entailment for claims in {W}ikipedia}.
\newblock In \emph{Proceedings of the 2023 Conference on Empirical Methods in
  Natural Language Processing}, pages 7561--7583, Singapore. Association for
  Computational Linguistics.

\bibitem[{Lin(2004)}]{lin2004rouge}
Chin-Yew Lin. 2004.
\newblock \href {https://aclanthology.org/W04-1013} {{ROUGE}: A package for
  automatic evaluation of summaries}.
\newblock In \emph{Text Summarization Branches Out}, pages 74--81, Barcelona,
  Spain. Association for Computational Linguistics.

\bibitem[{Liu et~al.(2023)Liu, Fabbri, Liu, Zhao, Nan, Han, Han, Joty, Wu,
  Xiong, and Radev}]{liu-etal-2023-revisiting}
Yixin Liu, Alex Fabbri, Pengfei Liu, Yilun Zhao, Linyong Nan, Ruilin Han,
  Simeng Han, Shafiq Joty, Chien-Sheng Wu, Caiming Xiong, and Dragomir Radev.
  2023.
\newblock \href {https://doi.org/10.18653/v1/2023.acl-long.228} {Revisiting the
  gold standard: Grounding summarization evaluation with robust human
  evaluation}.
\newblock In \emph{Proceedings of the 61st Annual Meeting of the Association
  for Computational Linguistics (Volume 1: Long Papers)}, pages 4140--4170,
  Toronto, Canada. Association for Computational Linguistics.

\bibitem[{Maheshwari et~al.(2024)Maheshwari, Bandyopadhyay, Garimella, and
  Natarajan}]{maheshwari2024presentations}
Himanshu Maheshwari, Sambaran Bandyopadhyay, Aparna Garimella, and Anandhavelu
  Natarajan. 2024.
\newblock \href {https://doi.org/10.48550/ARXIV.2405.13095} {Presentations are
  not always linear! {GNN} meets {LLM} for document-to-presentation
  transformation with attribution}.
\newblock \emph{CoRR}, arXiv:2405.13095.

\bibitem[{Mesnard et~al.(2024)Mesnard, Hardin, Dadashi, Bhupatiraju, Pathak,
  Sifre, Rivi{\`{e}}re, Kale, Love, Tafti, Hussenot, Chowdhery, Roberts, Barua,
  Botev, Castro{-}Ros, Slone, H{\'{e}}liou, Tacchetti, Bulanova, Paterson,
  Tsai, Shahriari, Lan, Choquette{-}Choo, Crepy, Cer, Ippolito, Reid,
  Buchatskaya, Ni, Noland, Yan, Tucker, Muraru, Rozhdestvenskiy, Michalewski,
  Tenney, Grishchenko, Austin, Keeling, Labanowski, Lespiau, Stanway, Brennan,
  Chen, Ferret, Chiu, and et~al.}]{team2024gemma}
Thomas Mesnard, Cassidy Hardin, Robert Dadashi, Surya Bhupatiraju, Shreya
  Pathak, Laurent Sifre, Morgane Rivi{\`{e}}re, Mihir~Sanjay Kale, Juliette
  Love, Pouya Tafti, L{\'{e}}onard Hussenot, Aakanksha Chowdhery, Adam Roberts,
  Aditya Barua, Alex Botev, Alex Castro{-}Ros, Ambrose Slone, Am{\'{e}}lie
  H{\'{e}}liou, Andrea Tacchetti, Anna Bulanova, Antonia Paterson, Beth Tsai,
  Bobak Shahriari, Charline~Le Lan, Christopher~A. Choquette{-}Choo,
  Cl{\'{e}}ment Crepy, Daniel Cer, Daphne Ippolito, David Reid, Elena
  Buchatskaya, Eric Ni, Eric Noland, Geng Yan, George Tucker, George{-}Cristian
  Muraru, Grigory Rozhdestvenskiy, Henryk Michalewski, Ian Tenney, Ivan
  Grishchenko, Jacob Austin, James Keeling, Jane Labanowski, Jean{-}Baptiste
  Lespiau, Jeff Stanway, Jenny Brennan, Jeremy Chen, Johan Ferret, Justin Chiu,
  and et~al. 2024.
\newblock \href {https://doi.org/10.48550/ARXIV.2403.08295} {Gemma: Open models
  based on gemini research and technology}.
\newblock \emph{CoRR}, abs/2403.08295.

\bibitem[{Min et~al.(2023)Min, Krishna, Lyu, Lewis, Yih, Koh, Iyyer,
  Zettlemoyer, and Hajishirzi}]{min2023factscore}
Sewon Min, Kalpesh Krishna, Xinxi Lyu, Mike Lewis, Wen-tau Yih, Pang Koh, Mohit
  Iyyer, Luke Zettlemoyer, and Hannaneh Hajishirzi. 2023.
\newblock \href {https://doi.org/10.18653/v1/2023.emnlp-main.741}
  {{FA}ct{S}core: Fine-grained atomic evaluation of factual precision in long
  form text generation}.
\newblock In \emph{Proceedings of the 2023 Conference on Empirical Methods in
  Natural Language Processing}, pages 12076--12100, Singapore. Association for
  Computational Linguistics.

\bibitem[{Nakano et~al.(2021)Nakano, Hilton, Balaji, Wu, Ouyang, Kim, Hesse,
  Jain, Kosaraju, Saunders, Jiang, Cobbe, Eloundou, Krueger, Button, Knight,
  Chess, and Schulman}]{nakano2021webgpt}
Reiichiro Nakano, Jacob Hilton, Suchir Balaji, Jeff Wu, Long Ouyang, Christina
  Kim, Christopher Hesse, Shantanu Jain, Vineet Kosaraju, William Saunders,
  Xu~Jiang, Karl Cobbe, Tyna Eloundou, Gretchen Krueger, Kevin Button, Matthew
  Knight, Benjamin Chess, and John Schulman. 2021.
\newblock \href {https://arxiv.org/abs/2112.09332} {Webgpt: Browser-assisted
  question-answering with human feedback}.
\newblock \emph{CoRR}, abs/2112.09332.

\bibitem[{Nenkova and
  Passonneau(2004)}]{nenkova-passonneau-2004-evaluating-pyramid}
Ani Nenkova and Rebecca Passonneau. 2004.
\newblock \href {https://aclanthology.org/N04-1019} {Evaluating content
  selection in summarization: The pyramid method}.
\newblock In \emph{Proceedings of the Human Language Technology Conference of
  the North {A}merican Chapter of the Association for Computational
  Linguistics: {HLT}-{NAACL} 2004}, pages 145--152, Boston, Massachusetts, USA.
  Association for Computational Linguistics.

\bibitem[{Ni et~al.(2019)Ni, Li, and McAuley}]{ni-etal-2019-justifying}
Jianmo Ni, Jiacheng Li, and Julian McAuley. 2019.
\newblock \href {https://doi.org/10.18653/v1/D19-1018} {Justifying
  recommendations using distantly-labeled reviews and fine-grained aspects}.
\newblock In \emph{Proceedings of the 2019 Conference on Empirical Methods in
  Natural Language Processing and the 9th International Joint Conference on
  Natural Language Processing (EMNLP-IJCNLP)}, pages 188--197, Hong Kong,
  China. Association for Computational Linguistics.

\bibitem[{Nie et~al.(2020)Nie, Williams, Dinan, Bansal, Weston, and
  Kiela}]{nie2020adversarial}
Yixin Nie, Adina Williams, Emily Dinan, Mohit Bansal, Jason Weston, and Douwe
  Kiela. 2020.
\newblock \href {https://doi.org/10.18653/v1/2020.acl-main.441} {Adversarial
  {NLI}: A new benchmark for natural language understanding}.
\newblock In \emph{Proceedings of the 58th Annual Meeting of the Association
  for Computational Linguistics}, pages 4885--4901, Online. Association for
  Computational Linguistics.

\bibitem[{Parsons(1990)}]{parsons1990events}
Terence Parsons. 1990.
\newblock \emph{Events in the Semantics of English: A Study in Subatomic
  Semantics}.
\newblock MIT Press.

\bibitem[{Pavlick and Kwiatkowski(2019)}]{pavlick-kwiatkowski-2019-inherent}
Ellie Pavlick and Tom Kwiatkowski. 2019.
\newblock \href {https://doi.org/10.1162/tacl_a_00293} {Inherent disagreements
  in human textual inferences}.
\newblock \emph{Transactions of the Association for Computational Linguistics},
  7:677--694.

\bibitem[{Raffel et~al.(2020)Raffel, Shazeer, Roberts, Lee, Narang, Matena,
  Zhou, Li, and Liu}]{raffel2020exploring}
Colin Raffel, Noam Shazeer, Adam Roberts, Katherine Lee, Sharan Narang, Michael
  Matena, Yanqi Zhou, Wei Li, and Peter~J. Liu. 2020.
\newblock \href {http://jmlr.org/papers/v21/20-074.html} {Exploring the limits
  of transfer learning with a unified text-to-text transformer}.
\newblock \emph{J. Mach. Learn. Res.}, 21:140:1--140:67.

\bibitem[{Russell(2014)}]{russell2009philosophy}
Bertrand Russell. 2014.
\newblock \href {https://doi.org/10.5840/monist19182843} {{The Philosophy of
  Logical Atomism.}}
\newblock \emph{The Monist}, 28(4):495--527.

\bibitem[{Scir{\`e} et~al.(2024)Scir{\`e}, Ghonim, and
  Navigli}]{scire-etal-2024-fenice}
Alessandro Scir{\`e}, Karim Ghonim, and Roberto Navigli. 2024.
\newblock \href {https://doi.org/10.18653/v1/2024.findings-acl.841} {{FENICE}:
  Factuality evaluation of summarization based on natural language inference
  and claim extraction}.
\newblock In \emph{Findings of the Association for Computational Linguistics
  ACL 2024}, pages 14148--14161, Bangkok, Thailand and virtual meeting.
  Association for Computational Linguistics.

\bibitem[{Stainton(2005)}]{stainton2005defense}
Robert~J. Stainton. 2005.
\newblock \href {https://doi.org/10.1093/acprof:oso/9780199251520.003.0011}
  {{383In Defense of Non-Sentential Assertion}}.
\newblock In \emph{{Semantics versus Pragmatics}}. Oxford University Press.

\bibitem[{Tiedemann and Mur(2008)}]{tiedemann2008simple}
J{\"o}rg Tiedemann and Jori Mur. 2008.
\newblock \href {https://aclanthology.org/W08-1803} {Simple is best:
  Experiments with different document segmentation strategies for passage
  retrieval}.
\newblock In \emph{Coling 2008: Proceedings of the 2nd workshop on Information
  Retrieval for Question Answering}, pages 17--25, Manchester, UK. Coling 2008
  Organizing Committee.

\bibitem[{Wanner et~al.(2024)Wanner, Ebner, Jiang, Dredze, and
  Durme}]{wanner2024closer}
Miriam Wanner, Seth Ebner, Zhengping Jiang, Mark Dredze, and Benjamin~Van
  Durme. 2024.
\newblock \href {https://doi.org/10.48550/ARXIV.2403.11903} {A closer look at
  claim decomposition}.
\newblock \emph{CoRR}, abs/2403.11903.

\bibitem[{Weber-Genzel et~al.(2024)Weber-Genzel, Peng, De~Marneffe, and
  Plank}]{weber-genzel-etal-2024-varierr}
Leon Weber-Genzel, Siyao Peng, Marie-Catherine De~Marneffe, and Barbara Plank.
  2024.
\newblock \href {https://doi.org/10.18653/v1/2024.acl-long.123} {{V}ari{E}rr
  {NLI}: Separating annotation error from human label variation}.
\newblock In \emph{Proceedings of the 62nd Annual Meeting of the Association
  for Computational Linguistics (Volume 1: Long Papers)}, pages 2256--2269,
  Bangkok, Thailand. Association for Computational Linguistics.

\bibitem[{Zhang et~al.(2023)Zhang, Elfardy, Dreyer, Small, Ji, and
  Bansal}]{zhang-etal-2023-enhancing}
Zixuan Zhang, Heba Elfardy, Markus Dreyer, Kevin Small, Heng Ji, and Mohit
  Bansal. 2023.
\newblock \href {https://doi.org/10.18653/v1/2023.eacl-main.124} {Enhancing
  multi-document summarization with cross-document graph-based information
  extraction}.
\newblock In \emph{Proceedings of the 17th Conference of the European Chapter
  of the Association for Computational Linguistics}, pages 1696--1707,
  Dubrovnik, Croatia. Association for Computational Linguistics.

\end{thebibliography}

\clearpage

\appendix

\section{\changeA{Metric Calculation Examples}}
\label{sec:metric_calc_examples}
We show examples of how metrics are calculated in Table \ref{tab:metric_calc_examples}. In each row, we show an example and the expected and calculated metrics for it. We also mention which propositions property is mainly measured by the metric. Finally, we provide an explanation about how the property affects the score.

\begin{table*}[ht!]
\small
\centering
\begin{tabular}{|p{2cm}|p{2cm}|p{3.5cm}|p{1.3cm}|p{1.3cm}|p{3cm}|}

\hline
\textbf{Metric} & \textbf{Property} & \textbf{Example}  & \textbf{Expected Score} & \textbf{Calculated Score} & \textbf{Explanation} \\ 
\hline
Reference-less precision & Supported & Passage = "The price of the books are all less than ten dollars, and they download before you can get up for a cup of coffee." -- Predicted Propositions = ["The books download before you can get up for a cup of coffee."] & 1 & 0.9999 & The predicted proposition is entailed by the passage. \\
\hline
Reference-less recall & Comprehensive & Passage = "The price of the books are all less than ten dollars, and they download before you can get up for a cup of coffee." -- Predicted Propositions = ["The price of the books are all less than ten dollars."] & 0 & 0 & The predicted propositions do not cover all the information in the passage. \\
\hline
Reference-based precision & Self-contained & Predicted Propositions = ["The price of the books are all less than ten dollars.", "They download before you can get up for a cup of coffee."] -- Gold Propositions = ["The price of the books are all less than ten dollars.", "The books download before you can get up for a cup of coffee."] & 0.5 & 0.5 & The second predicted proposition is not as self-contained as the second gold proposition ("They" vs "The books"). Therefore, the second predicted proposition should get a score of 0 when calculating reference-based precision. \\
\hline
Reference-based recall & Self-contained & Predicted Propositions = ["The price of the books are all less than ten dollars.", "They download before you can get up for a cup of coffee."] -- Gold Propositions = ["The price of the books are all less than ten dollars.", "The books download before you can get up for a cup of coffee."] & 0.5 & 0.5 & The second gold proposition is more self-contained than the second predicted proposition ("The books" vs "They"). Therefore, the second gold proposition should get a score of 0 when calculating reference-based recall. \\
\hline
Reference-based precision & Atomic & Predicted Propositions = ["The price of the books are all less than ten dollars. They download before you can get up for a cup of coffee."] -- Gold Propositions = ["The price of the books are all less than ten dollars.", "The books download before you can get up for a cup of coffee."] & 0 & 0 & The predicted proposition is not as atomic as any of the gold proposition. Therefore, it gets a score of 0 when calculating reference-based precision. \\
\hline
Reference-based recall & Atomic & Predicted Propositions = ["The price of the books are all less than ten dollars. They download before you can get up for a cup of coffee."] -- Gold Propositions = ["The price of the books are all less than ten dollars.", "The books download before you can get up for a cup of coffee."] & 0 & 0 & The gold propositions are more atomic than the predicted proposition. Therefore, they both get a score of 0 when calculating reference-based recall. \\
\hline
\end{tabular}
\caption{Examples with expected and calculated metrics. For each example, we provide the propositions property that is mainly measured by the metric. In addition, we explain how the property affects the score.}
\label{tab:metric_calc_examples}
\end{table*}

\section{Few-shot Prompting Example for Synthetic Multi-Domain Text Generation}
\label{sec:few-shot-example}

Table \ref{tab:few-shot-example} shows one of the $18$ few-shot examples used to generate synthetic multi-domain data (Section \ref{sec:syn-data}). The full list can be found in \citet{hosseini2024synthetic}.

\begin{table*}[ht!]
\small
\centering
\begin{tabular}{| m{1.5cm} | m{1.5cm} | m{10cm} |}
\hline
\textbf{Domain} & \textbf{Length} & \textbf{Text} \\
\hline
reddit post & paragraph & Hey there everyone! I often see people asking where to start when getting into prog metal, so I thought instead of answering every one of them individually I'd make a list. I'm not going into too much depth because otherwise this will become endless, but I'll try to give a brief explanation of all styles I'm going over. So let's get started! \\
\hline
\end{tabular}
\caption{A few-shot example used to generate synthetic multi-domain text. The example has a domain, a length, and a text.}
\label{tab:few-shot-example}
\end{table*}

\section{Hyper-parameters}
\label{sec:hparam}

For training Gemma modals, we used a batch size of $8$, and an initial learning rate of $5e-5$ and minimum learning rate of $5e-7$ with linear warmup cosine annealing (warmup step of $100$ and cosine decay exp $1.0$). We trained for $1$ epoch.

For all few-shot models, we used a temperature of $0$. We tried higher temperatures, but the results were worse.

We trained Gemini Pro with two different learning rates, $1e-4$ and $1e-5$, and selected the first one since it gave better results on ROSE development set. We trained the model with a batch size of $32$ for around $4$ epochs. We saved checkpoints every $50$ steps and selected the one with lowest loss on the development set.

\section{Pre-processing the ROSE Dataset, Aligning Propositions with Sentences, and Filtering Problematic Examples}
\label{sec:alignment_and_filter}

\changeA{We pre-process the dataset to improve its quality based on the following steps:}

\changeA{A) In some cases, ACUs end with a space before the period. We remove the extra space. Additionally, some ACUs do not end with a period (and do not end with ``...'' either). In these cases, we add a period to the ACU.}

\changeA{B) For each sentence, the ROSE dataset annotators first write an ACU consisting the main information from the subject of the main clause. Then, they add one ACU for each additional information in the sentence by adding minimal necessary information to the original ACU \cite{liu-etal-2023-revisiting}. In some cases, the original ACU is exactly repeated in other ACUs. In these cases, we removed the first ACU as they are often very short and not very informative. For example, the ACU ``Many seals are shot'' is removed because we also have another ACU ``Many seals are shot to death for their fur''.}

C) As explained in \S\ref{sec:datasets}, we align propositions in the ROSE dataset with their corresponding sentences, and filter problematic examples. In particular, we follow these steps:

\begin{enumerate}[itemsep=-1ex]
    \item For each proposition $j$:
        \begin{enumerate}
        \item compute NLI (sentence $i$, proposition $j$) for all sentences. If the sentence with maximum NLI score to proposition $j$ has a score $\geq \tau{=}0.9$, then we use that sentence as the alignment. Otherwise:
        \item Compute NLI (prefix (${i-1}) +$  sentence $i$) for all sentences, where prefix ($i-1$) means the sentences up to sentence $i$, and $+$ means space concatenation. Find the first sentence which yields entailment score $\geq \tau{=}0.9$ (if any). If such a sentence exists, we use that as the alignment. Otherwise, we discard the example.
        \end{enumerate}
    \item If a sentence is not aligned with any proposition, we discard the whole example.
\end{enumerate}

The reason that in step 1 (b), we add the prefix of the sentences before computing the NLI score is that sometimes the full context is necessary to obtain a high NLI score, e.g., cases where the sentence contains a pronoun, but the proposition has the full name. After aligning all the propositions with the above approach, we autoamtically remove examples that have unsupported propositions, and cases where a sentence might not have any propositions, a special case of non-comprehensive propositions.

Table \ref{tab:filter_unsupported} and Table \ref{tab:filter_not_comprehensive} show filtered examples with unsupported propositions and non-comprehensive propositions list, respectively.

\begin{table*}[ht!]
\small
    \centering
    \begin{tabular}{|>{\raggedright\arraybackslash}p{15cm}|}
        \hline
        \multicolumn{1}{|c|}{ {\bf \textsc{ Input Text } } } \\ \hline
        Packs of wild boar are hunting newborn lambs in Britain, experts claim. \textbf{Boar at the Forest of Dean usually feed only on plants and dead animals.} But in recent weeks, groups of boar have reportedly killed four lambs. Serious implications for animal health and spread of disease, vet says. \\ \hline
        \multicolumn{1}{|c|}{ {\bf \textsc{ Propositions } } } \\ \hline
        \begin{itemize}[itemsep=-1ex]
            \item newborn lambs are hunted.
            \item Packs of wild boar are hunting in Britain.
            \item Packs of wild boar are hunting, experts claim.
            \item \textbf{Boar usually feed only on plants.}
            \item \textbf{Boar usually feed only on dead animals.}
            \item The boar is from the Forest of Dean.
            \item groups of boar have reportedly killed lambs.
            \item In recent weeks,  four lambs are killed.
            \item Serious implications for animal health.
            \item Serious implications for spread of disease.
            \item They are serious implications, vet says.
        \end{itemize}
        \\ \hline
    \end{tabular}
    \caption{Example from the ROSE dataset where propositions are not supported by the input text, but we filter the example out. The relevant sentence and unsupported propositions is boldfaced.}
    \label{tab:filter_unsupported}
\end{table*}

\begin{table*}[ht!]
\small
    \centering
    \begin{tabular}{|>{\raggedright\arraybackslash}p{15cm}|}
        \hline
        \multicolumn{1}{|c|}{ {\bf \textsc{ Input Text } } } \\ \hline
        Wembley was almost full for England's 4-0 win over Lithunia. \textbf{Raheem Sterling linked well with Wayne Rooney and Danny Welbeck. Roy Hodgson must prepare his side for the stiffer tests at Euro 2016. Italy are a different proposition to the side that beat England last summer.} \\ \hline
        \multicolumn{1}{|c|}{ {\bf \textsc{ Propositions } } } \\ \hline
        \begin{itemize}[itemsep=-1ex]
            \item Wembley was almost full.
            \item England won.
            \item The score was 4-0.
            \item England played Lithuania.
        \end{itemize}
        \\ \hline
    \end{tabular}
    \caption{Example from the ROSE dataset where propositions are not comprehensive, but we filter the example out. The sentences that are not covered by propositions are boldfaced.}
    \label{tab:filter_not_comprehensive}
\end{table*}

\section{Full Results on All Datasets}
\label{sec:full-results}

In this section, we show all the results reported in \S\ref{sec:exp} plus two additional sets of results. Table \ref{tab:res_rose_full}, \ref{tab:res_reddit_full}, and \ref{tab:res_amazon_full} show the full results.

\textit{\textbf{Few-shot results with random examples and examples from \citet{wanner2024closer}}}. In \S\ref{sec:exp}, we showed few-shot prompting results with dynamically selected examples. In this section, we also add few-shot prompting results with few-shot examples randomly selected per test example. We performed the experiment 5 times. In most cases, the dynmaic approach outperforms the random approach. This is expected since the LLM can learn more from more similar few-shot examples than random examples.

We also experimented with $21$ few-shot prompts from \citet{wanner2024closer}. These examples are annotated based on Bertrand Russell’s theory of logical atomism \cite{russell2009philosophy} and neo Davidsonian analysis \cite{davidson1967logical,parsons1990events}. This few-shot prompting approach led to generally worse results on all datasets and all metrics, including $RL_r$ (with only one exception).

\textit{\textbf{Gemma 2B results}}. In \S\ref{sec:exp}, we trained Gemma 7B on ROSE and also trained it as a student on distillation data. In this section, we additionally report the results with Gemma 2B. Gemma 2B generally performs slightly worse than Gemma 7B, but we obtain the same trends for Gemma 2B as Gemma 7B. For example, Gemma 2B student models obtain similar results to teacher models and generally obtain better results than Gemma 2B trained on ROSE.

\begin{table*}[ht!] 
\centering 
\small 
\begin{tabular}{|m{0.16\linewidth}|*{3}{m{0.06\linewidth}}|*{3}{m{0.06\linewidth}}|m{0.06\linewidth}|} 
& \multicolumn{3}{c}{\bf \textsc{Reference-Less Metrics}} & \multicolumn{3}{c}{\bf \textsc{Reference-Based Metrics}} & \\& Precision & Recall & F1 & Precision & Recall & F1 & \# Props \\
\hline
Gold & 99.71 & 96.54 & 97.53 & 100.00 & 100.00 & 100.00 & 5.84 \\ \hline
Sentence & 100.00 & 100.00 & 100.00 & 24.71 & 18.24 & 20.42 & 2.52 \\ \hline
\multicolumn{8}{|c|}{ {\bf \textsc{ Few Shot } } } \\ \hline
Gemini Pro Ran & 94.58 & 93.11 & 91.05 & 45.96 & 48.25 & 45.49 & 6.01 \\ \hline
Gemini Pro R-ND & 94.63 & 90.43 & 89.48 & 40.08 & 50.82 & 42.58 & 8.00 \\ \hline
Gemini Pro Dyn & 99.21 & 93.26 & 94.31 & 47.10 & 41.41 & 43.20 & 4.22 \\ \hline
Gemini Ultra Ran & 89.10 & 91.01 & 85.89 & 44.86 & 50.71 & 45.58 & 8.39 \\ \hline
Gemini Ultra R-ND & 97.70 & 89.55 & 90.92 & 33.97 & 51.24 & 39.72 & 8.93 \\ \hline
Gemini Ultra Dyn & 99.37 & 89.75 & 91.88 & 49.49 & 47.49 & 47.74 & 5.29 \\ \hline
\multicolumn{8}{|c|}{ {\bf \textsc{ Trained on ROSE } } } \\ \hline
Gemma 2B UG & 96.49 & 92.64 & 92.67 & 51.75 & 49.75 & 50.20 & 5.56 \\ \hline
Gemma 2B G & 97.46 & 94.49 & 94.39 & 53.29 & 51.59 & 51.89 & 5.62 \\ \hline
Gemma 7B UG & 98.09 & 96.57 & 96.54 & 52.16 & 50.93 & 51.02 & 5.57 \\ \hline
Gemma 7B G & 98.57 & 97.48 & 97.48 & 53.70 & 51.43 & 51.93 & 5.61 \\ \hline
Gemini Pro UG & 99.51 & 97.84 & 98.20 & 54.76 & 52.48 & 53.02 & 5.54 \\ \hline
Gemini Pro G & 99.31 & 96.66 & 97.23 & 55.96 & 54.87 & 54.83 & 5.66 \\ \hline
Gemini Ultra UG & 99.46 & 98.05 & 98.33 & \textbf{57.69} & 56.32 & 56.45 & 5.72 \\ \hline
Gemini Ultra G & \textbf{99.53} & \textbf{98.16} & \textbf{98.50} & 57.62 & \textbf{56.50} & \textbf{56.49} & \textbf{5.77} \\ \hline
\multicolumn{8}{|c|}{ {\bf \textsc{ Gemma 2B Distilled Models } } } \\ \hline
Gemini Pro Data & 98.20 & 96.40 & 96.61 & 54.13 & 52.29 & 52.61 & 5.46 \\ \hline
Gemini Ultra Data & 97.55 & 97.31 & 96.92 & 54.73 & 53.04 & 53.30 & 5.64 \\ \hline
\multicolumn{8}{|c|}{ {\bf \textsc{ Gemma 7B Distilled Models } } } \\ \hline
Gemini Pro Data & 98.98 & 97.91 & 98.02 & 55.14 & 53.02 & 53.50 & 5.53 \\ \hline
Gemini Ultra Data & 98.93 & 98.08 & 98.23 & 56.82 & 55.18 & 55.41 & 5.65 \\ \hline
\end{tabular}
\caption{Full results on the \textsc{ROSE} dataset. The methods are split into $5$ blocks. The first block has gold and sentence baselines. The second one has few-shot baselines with randomly (Ran) selected examples, examples from \citet{wanner2024closer} based on based on Russellian and neo-Davidsonian theories (R-ND), and dynamically (Dyn) selected examples. The third block has baselines directly trained on ROSE with ungrouped (UG) and grouped (G) propositions. The fourth and fifth blocks contain Gemma 7B and Gemma 2B distilled models results, respectively. The best result for each metric (excluding gold and sentence baselines) are boldfaced.}
\label{tab:res_rose_full}
\end{table*}

\begin{table*}[ht!] 
\centering 
\small 
\begin{tabular}{|m{0.16\linewidth}|*{3}{m{0.06\linewidth}}|*{3}{m{0.06\linewidth}}|m{0.06\linewidth}|} 
& \multicolumn{3}{c}{\bf \textsc{Reference-Less Metrics}} & \multicolumn{3}{c}{\bf \textsc{Reference-Based Metrics}} & \\& Precision & Recall & F1 & Precision & Recall & F1 & \# Props \\
\hline
Gold & 98.72 & 99.22 & 98.86 & 100.00 & 100.00 & 100.00 & 10.70 \\ \hline
Sentence & 100.00 & 100.00 & 100.00 & 32.99 & 17.80 & 22.43 & 4.40 \\ \hline
\multicolumn{8}{|c|}{ {\bf \textsc{ Few Shot } } } \\ \hline
Gemini Pro Ran & 97.06 & 80.48 & 84.36 & 54.10 & 47.24 & 49.79 & 8.92 \\ \hline
Gemini Pro R-ND & 94.27 & 75.33 & 81.57 & 54.83 & \textbf{50.98} & \textbf{51.98} & 11.60 \\ \hline
Gemini Pro Dyn & \textbf{100.00} & 83.40 & 89.31 & \textbf{56.98} & 42.03 & 47.74 & 7.30 \\ \hline
Gemini Ultra Ran & 97.87 & 71.44 & 78.37 & 48.69 & 43.35 & 45.04 & 9.04 \\ \hline
Gemini Ultra R-ND & 96.48 & 64.60 & 73.07 & 43.51 & 41.70 & 42.00 & 11.40 \\ \hline
Gemini Ultra Dyn & 98.99 & 72.61 & 80.44 & 54.59 & 44.73 & 48.53 & 8.15 \\ \hline
\multicolumn{8}{|c|}{ {\bf \textsc{ Trained on ROSE } } } \\ \hline
Gemma 2B G & 93.95 & 97.49 & 95.30 & 33.78 & 32.22 & 32.49 & 10.40 \\ \hline
Gemma 7B G & 98.25 & 98.73 & 98.35 & 35.49 & 38.30 & 36.57 & 10.25 \\ \hline
Gemini Pro G & 98.80 & 94.41 & 96.08 & 41.80 & 43.44 & 42.06 & \textbf{10.65} \\ \hline
Gemini Ultra G & 99.68 & 96.66 & 97.83 & 40.82 & 44.69 & 42.39 & 11.20 \\ \hline
\multicolumn{8}{|c|}{ {\bf \textsc{ Gemma 2B Distilled Models } } } \\ \hline
Gemini Pro Data & 98.85 & 97.08 & 97.87 & 46.78 & 46.07 & 45.57 & 11.00 \\ \hline
Gemini Ultra Data & 99.54 & 99.71 & \textbf{99.61} & 40.00 & 42.36 & 40.84 & 10.80 \\ \hline
\multicolumn{8}{|c|}{ {\bf \textsc{ Gemma 7B Distilled Models } } } \\ \hline
Gemini Pro Data & 99.47 & 97.08 & 98.00 & 45.22 & 48.08 & 46.20 & 10.90 \\ \hline
Gemini Ultra Data & 98.88 & \textbf{99.96} & 99.40 & 40.43 & 43.21 & 41.46 & 11.00 \\ \hline
\end{tabular}
\caption{Full results on the \textsc{REDDIT} dataset. See Table \ref{tab:res_rose_full}'s caption for details.}
\label{tab:res_reddit_full}
\end{table*}

\begin{table*}[ht!] 
\centering 
\small 
\begin{tabular}{|m{0.16\linewidth}|*{3}{m{0.06\linewidth}}|*{3}{m{0.06\linewidth}}|m{0.06\linewidth}|} 
& \multicolumn{3}{c}{\bf \textsc{Reference-Less Metrics}} & \multicolumn{3}{c}{\bf \textsc{Reference-Based Metrics}} & \\& Precision & Recall & F1 & Precision & Recall & F1 & \# Props \\
\hline
Gold & 100.00 & 99.98 & 99.99 & 100.00 & 100.00 & 100.00 & 6.55 \\ \hline
Sentence & 100.00 & 100.00 & 100.00 & 37.94 & 24.50 & 29.21 & 3.55 \\ \hline
\multicolumn{8}{|c|}{ {\bf \textsc{ Few Shot } } } \\ \hline
Gemini Pro Ran & 99.14 & 66.64 & 74.33 & 46.51 & 43.43 & 44.50 & 6.09 \\ \hline
Gemini Pro R-ND & 97.48 & 55.97 & 65.97 & 38.90 & 43.89 & 40.80 & 7.75 \\ \hline
Gemini Pro Dyn & 99.54 & 62.97 & 71.82 & 50.02 & 46.74 & 47.89 & 6.10 \\ \hline
Gemini Ultra Ran & 95.80 & 49.00 & 57.19 & 42.62 & 39.81 & 40.22 & 6.82 \\ \hline
Gemini Ultra R-ND & 98.75 & 44.38 & 56.02 & 29.99 & 36.42 & 32.47 & 8.45 \\ \hline
Gemini Ultra Dyn & 85.09 & 53.10 & 57.17 & 44.41 & 44.38 & 42.06 & 11.60 \\ \hline
\multicolumn{8}{|c|}{ {\bf \textsc{ Trained on ROSE } } } \\ \hline
Gemma 2B G & 97.60 & 98.98 & 98.01 & 50.89 & 49.53 & 49.99 & 6.35 \\ \hline
Gemma 7B G & \textbf{99.98} & \textbf{99.99} & \textbf{99.99} & 51.90 & 48.88 & 49.83 & 6.25 \\ \hline
Gemini Pro G & 99.53 & 97.98 & 98.50 & 55.62 & 54.09 & 54.52 & 6.60 \\ \hline
Gemini Ultra G & 99.83 & 96.98 & 98.09 & \textbf{56.75} & 57.08 & \textbf{56.65} & 6.80 \\ \hline
\multicolumn{8}{|c|}{ {\bf \textsc{ Gemma 2B Distilled Models } } } \\ \hline
Gemini Pro Data & 98.99 & 94.72 & 96.55 & 56.29 & 56.87 & 56.35 & 6.70 \\ \hline
Gemini Ultra Data & 99.77 & 96.30 & 97.77 & 54.19 & 53.97 & 53.89 & 6.60 \\ \hline
\multicolumn{8}{|c|}{ {\bf \textsc{ Gemma 7B Distilled Models } } } \\ \hline
Gemini Pro Data & 98.30 & 96.72 & 97.30 & 56.00 & \textbf{57.09} & 56.25 & 7.05 \\ \hline
Gemini Ultra Data & 99.27 & 99.16 & 99.21 & 53.43 & 53.03 & 53.09 & \textbf{6.55} \\ \hline
\end{tabular}
\caption{Full results on the \textsc{AMAZON Review} dataset. See Table \ref{tab:res_rose_full}'s caption for details.}
\label{tab:res_amazon_full}
\end{table*}

\section{Few-shot Models Recall Issues}
\label{sec:fewshot_vs_student}

Table \ref{tab:fewshot_vs_student} shows examples where a few-shot model (Gemini Pro with dynamically selected examples) does not cover some of the facts from the input text, but our student model (Gemma 7B distilled from Gemini Pro data) successfully covers those facts.

\begin{table*}[ht!]
\small
\centering
\begin{tabular}{| m{3.7cm} | m{5.3cm} | m{5.3cm} |}
\hline
\textbf{Input text} & \textbf{Few-shot} & \textbf{Gemma 7B distilled from Gemini Pro} \\
\hline
\textbf{In the 1940's}, the NAACP charged the film with giving the impression of "an idyllic master-slave relationship." & - The NAACP charged the film with giving the impression of "an idyllic master-slave relationship." &  - In the 1940's, the NAACP charged the film. - the NAACP charged the film with giving the impression of "an idyllic master-slave relationship." \\ \hline
The scope of such restrictions and requirements varies widely \textbf{from state to state and business to business}, but in almost all cases there are a ton of boxes which need to be checked before anyone can move forward. & - The scope of restrictions and requirements varies widely. - There are a ton of boxes which need to be checked before anyone can move forward. & - The scope of governmental restrictions varies widely from state to state. - The scope of governmental requirements varies widely from state to state. - The scope of governmental restrictions varies widely from business to business. - The scope of governmental requirements varies widely from business to business. - In almost all cases there are a ton of boxes which need to be checked. - There are a ton of boxes which need to be checked before anyone can move forward. \\ \hline
\end{tabular}
\caption{Examples where a few-shot model (Gemini Pro with dynamically selected examples) does not cover some the facts (boldfaced) from a sentence. However, our student model (Gemma 7B distilled from Gemini Pro data) covers those facts.}
\label{tab:fewshot_vs_student}
\vspace{-4mm}
\end{table*}

\end{document}